\documentclass[conference]{IEEEtran}
\usepackage{cite}
\usepackage{amsmath,amssymb,amsfonts}
\usepackage{algorithmic}
\usepackage{graphicx}
\usepackage{textcomp}
\usepackage{xcolor}
\usepackage{caption}
\usepackage{subcaption}
\usepackage{amsmath}
\usepackage{hyperref}
\usepackage[font=footnotesize,labelfont=bf]{caption}

\DeclareMathOperator{\abs}{abs}

\DeclareMathOperator{\sign}{sign}

\def\BibTeX{{\rm B\kern-.05em{\sc i\kern-.025em b}\kern-.08em
    T\kern-.1667em\lower.7ex\hbox{E}\kern-.125emX}}

\begin{document}

\title{Learning Controllable Content Generators}

\makeatletter
\newcommand{\linebreakand}{%
  \end{@IEEEauthorhalign}
  \hfill\mbox{}\par
  \mbox{}\hfill\begin{@IEEEauthorhalign}
}
\makeatother

\author{\IEEEauthorblockN{Sam Earle}
\IEEEauthorblockA{
\textit{New York University}\\
Brooklyn, New York \\
sam.earle@nyu.edu}
\and
\IEEEauthorblockN{Maria Edwards}
\IEEEauthorblockA{\textit{New York University} \\
Brooklyn, New York \\
mariaedwards@nyu.edu}
\and
\IEEEauthorblockN{Ahmed Khalifa}
\IEEEauthorblockA{
\textit{New York University} \\
Brooklyn, New York \\
ahmed@akhalifa.com}

\linebreakand
\IEEEauthorblockN{Philip Bontrager}
\IEEEauthorblockA{
\textit{TheTake} \\
New York, New York \\
pbontrager@gmail.com}
\and
\IEEEauthorblockN{Julian Togelius}
\IEEEauthorblockA{
\textit{New York University} \\
Brooklyn, New York \\
julian@togelius.com}
}

\maketitle

\begin{abstract}
It has recently been shown that reinforcement learning can be used to train generators capable of producing high-quality game levels, with quality defined in terms of some user-specified heuristic.
To ensure that these generators' output is sufficiently diverse (that is, not amounting to the reproduction of a single optimal level configuration), the generation process is constrained such that the initial seed results in some variance in the generator's output.
However, this results in a loss of control over the generated content for the human user.
We propose to train generators capable of producing controllably diverse output, by making them ``goal-aware.''
To this end, we add conditional inputs representing how close a generator is to some heuristic, and also modify the reward mechanism to incorporate that value.
Testing on multiple domains, we show that the resulting level generators are capable of exploring the space of possible levels in a targeted, controllable manner, producing levels of comparable quality as their goal-unaware counterparts, that are diverse along designer-specified dimensions.
\end{abstract}

\begin{IEEEkeywords}
procedural content generation, reinforcement learning, game AI, conditional generation, pcgrl, conditional generation
\end{IEEEkeywords}

\section{Introduction}

\begin{figure}
\begin{subfigure}{0.5\textwidth}
    \includegraphics[width=0.95\textwidth]{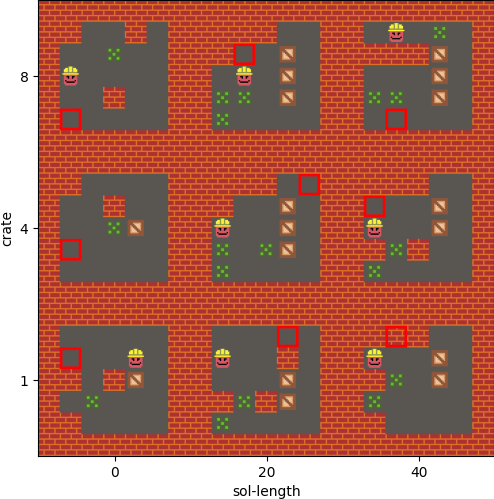}
\end{subfigure}
\caption{Generated levels from a trained generator on the game of Sokoban. The generator is controlled during inference to produce small sokoban levels with variable solution length and numbers of crates.}
\label{fig:sok_crate_sol_intro}
\end{figure}

The idea of using reinforcement learning to learn game content generators--or at least, the successful implementation of this idea--is relatively recent~\cite{khalifa2020pcgrl}. The basic idea is simple: an agent is trained to construct levels (or other types of game content) in the same way an agent would be trained to play the game. Instead of being rewarded for e.g. winning the game, the agent is rewarded for improving the level. 

Compared to using search or optimization methods for content generation~\cite{togelius2011search}, Procedural Content Generation via Reinforcement Learning (PCGRL) requires a long training time, but is then able to produce an arbitrary number of content artefacts very rapidly. Compared to methods based on supervised learning~\cite{summerville2018procedural,liu2020deep}, PCGRL avoids the need for training data, but instead requires a reward function that reflects content quality.

As PCGRL was only proposed as a methodology very recently, there are many issues that have not been studied yet. One of them is controllability. It is highly desirable for a content generation method to be able to be controlled by the user (such as a human designer or a difficulty adjustment algorithm). For example, one might want to develop a level that favors a specific playstyle, a quest with a certain degree of branching, or a map with a certain balance. The obvious way of controlling the output of an RL-trained content generator would be to change the reward function (just like how you would change the fitness function in search-based PCG), but that would mean retraining the generator for every change of control parameters.

In this paper, we explore approaches to training generators which are controllable \emph{after generation}. Or in other words, training single content generator agents to output a variety of content artefacts depending on control parameters. We do this by introducing control parameters as additional ``conditional'' inputs to the neural network, and then rewarding the generator not only for creating correct artefacts but also adhering to the control parameters. Figure~\ref{fig:sok_crate_sol_intro} shows an example of generated sokoban level by one of our trained generators. The generator is able to control the number of crates in the level and the solution length to provide us with different levels across these two dimensions.

The key to making this work is the control regime. To ensure the agent learns to respond to controls, we need to sample from the space of all possible values these controls may take (e.g. them a uniform random distribution). But as this space becomes large, this naive approach may become inefficient. For the approach to scale to complex controls, the control regime should focus on targets with the most learning potential for the agent.

Using a curiosity-based teacher algorithm to sample control targets, and the observation and reward scheme described above, we demonstrate controllable PCGRL in three simple 2D level-design domains as well as in free-play modes in SimCity and Rollercoaster Tycoon.\footnote{Code is available at \href{https://github.com/smearle/gym-pcgrl}{https://github.com/smearle/gym-pcgrl}}

\section{Related work}

\subsection{PCGRL}

The idea behind PCGRL is to frame the PCG process as a game of level design for the RL agent \cite{khalifa2020pcgrl}. The notion of building levels in a sequential, goal-driven way complements our intuition about the human level design process, though it is not the approach usually employed by machine learning systems leveraged for PCG \cite{volz2018evolving, bontrager2020fully, summerville2018procedural, liu2020deep}. In PCGRL, the agent is given a random level and can then, turn-by-turn, change tiles in the level as it sees fit. For each change that improves the quality of the level, it is given a reward, and it is penalized when it makes things worse. This approach allows a user with no access to training data to train a functional content generator, by having them instead design a reward function from which an agent can learn. Where ML-based PCG often struggles to produce feasible content, here, functional constraints are communicated to the generator during training via its reward. At the same time, we get the benefits of a neural network-based generator, which can generalize to unseen input to produce novel designs \cite{justesen2018illuminating}. This allows us to combine strengths of both search-based, and ML-based PCG.

In a naive implementation, PCGRL is susceptible to a particular kind of overfitting in which the agent learns to generate the same level every time. We want to avoid this, as we want to learn generators rather than levels.
To make the RL agent learn to generate a variety of levels, the original PCGRL paper employed a strategy to make the agent responsive to the initial random level provided at the beginning of each episode. This strategy was designed to limit the number of changes that the agent was allowed to make to the initial map before the level-generation episode terminated (i.e. ``game over'') during training. The threshold could be set very low, to make the agent very responsive to the provided starting state, or it could be made very high, to give the agent freedom to build a more optimal design. The philosophy behind this strategy was that diversity of agent behavior should be incentivized by the environment, rather than built into the RL model or update rule, so as to remain compatible with any RL training regime. In that same spirit, our approach here is to make PCG Agents controllable by a human designer through changes in the PCGRL framework itself, and not through custom RL algorithms.

Taking advantage of the responsive nature of a PCG Agent, there has been followup work to the original PCG paper that uses these trained agents as collaborative designers with a human designer \cite{delarosa2020mixed}. By training the agent with a limited number of allowed changes the agent learns to make very efficient changes which is ideal for working with a human designer where the human and agent take turns. The limitation in this scenario is that the designer cannot communicate intent to the agent and can only communicate through changes made in the environment. In this work we want to make the agent responsive to provided, explicit, design goals, which will hopefully allow for for more meaningful human-AI collaboration.

Contemporaneous with our previous work is the work by \cite{dennis2020emergent} and \cite{gisslen2021adversarial}. In \cite{dennis2020emergent} the authors devise a game between two game-playing RL agents and a level design agent. This regret minimization game gives the level generator the task of generating levels that are not too hard or too easy for the playing agents and thus removes the need for explicitly designing a level design reward function. In \cite{gisslen2021adversarial}, the authors also pair a playing and generating agents together to remove the need for an explicit reward function but adapt the level design to the agent mid-game. These approaches are out of the scope of this current work but display a direction for how our work can be adapted to remove the need for a designer to create a reward function.


\subsection{Controllable RL}

There are several possible approaches to training controllable RL agents. We could train the RL agent to respond behaviorally to text commands \cite{hill2020human}. We could train it with different policy networks \cite{mossalam2016multi} or memory modules \cite{mirowski2018learning} corresponding different goals. Or we could train it with conditional inputs and reward shaping, like in \cite{gisslen2021adversarial}, where level designer-agents observe an auxiliary input that corresponds to intended difficulty, and are rewarded for controlling a player agent's level of success on the generated levels.

All of these approaches could potentially be applied to PCGRL (though they have not, yet) and each could address different use cases. In this work we propose an algorithm-agnostic approach involving conditional inputs and reward-shaping. By encoding the goal into the state representation of the level, we allow any RL algorithm to be used and the goal information can be coded with the reward function, without requiring any extra data.

\subsection{Absolute learning progress teacher algorithm}

In the BipedalWalker environment, a single RL agent can be trained to navigate diverse terrain, by parameterizing the terrain-generation process with a continuous variable, and sampling this variable from a Gaussian Mixture Model that is fitted to the agent's Absolute Learning Progress during training (ALP-GMM)~\cite{portelas2020teacher}. The agent's reward is the same across all sub-tasks. The same ALP-GMM teacher algorithm is used in our work to generate a curriculum of content-generation sub-tasks: rather than sampling environments, it samples target behavior (i.e. level characteristics), each of which correspond to a particular conditional input scheme and reward function.

\section{Domains}

\subsection{Binary, Zelda, and Sokoban}

Our main experimental domains are the same ones as in~\cite{khalifa2020pcgrl} (where they are also described in more detail): Binary, Zelda, and Sokoban. In the Binary domain, tiles can be either wall or floor. The goal is to create the longest shortest path between any two tiles. Zelda is the GVGAI~\cite{perez2019general} game which is inspired by the dungeon system in the original Legend of Zelda game. Levels need to allow the player character to get the key and open the door, and can place enemies to fight in the level. Sokoban, finally, is the classic box-pushing puzzle. A solvable Sokoban level allows for a way in which the play can push each box onto a target.

\subsection{SimCity}

In the SimCity reinforcement learning environment based on the open-sourced SimCity 1 \cite{earle2020using} \cite{simcity}, the agent constructs a city by placing zones (residential, commercial, and industrial), infrastructure (road, rail, electricity lines, airports, harbors), and services (police and fire stations, and parks). Citizens, simulated by cellular automata, will inhabit and develop zones according to local desirability rules and global demand by zone-type, and similarly travel along roads, with flows determined by local rules and a fixed commute cycle.

RL has previously been used to generate player agents that can successfully maximize fixed reward functions comprising weighted combinations of different types of population. The agents here, on the other hand, are trained to build cities resulting in a range of population levels. This is a more plausible approach to generating general playing agents for this type of management sim game, in which artefacts (like cities or theme parks) may be functionally optimal in a multitude of ways. 

\subsection{micro-RCT}

In the minimal theme-park management reinforcement learning environment \textit{micro-rct}~\footnote{\href{https://github.com/smearle/micro-rct}{https://github.com/smearle/micro-rct}}, based on RollerCoaster Tycoon \cite{rct} and reproducing some of the implementation logic of OpenRCT2~\footnote{\href{https://github.com/OpenRCT2/OpenRCT2}{https://github.com/Open/RCT2/OpenRCT2}}, 
the agent builds an amusement park by adding attractions including concessions (food, drink, or gift shop), rides (both smaller gentle or thrill rides, like spiral slides and twisters; and larger coasters), other services (washrooms and first aid stalls) and paths (connects different areas in the park), that allows a fixed number of guests (with varying intensity and nausea tolerance thresholds) to frequent them and produce income for the park. Guests' mood is affected by their experience on rides (which will be determined by the ride's characteristics and the guests' preferences and tolerance levels) and internal needs such as hunger, thirst, and bladder. The game could be played with the goal of maximizing park income, or producing widespread (dis)satisfaction among park guests.

\section{Method}

A controllable RL level generator is trained by feeding the generator inputs corresponding to target level features, then assigning reward to the generator when it produces levels with these features. New valued for target features (e.g., number of disjoint regions or length of longest path) are sampled uniformly at the beginning of each episode. The episode is terminated when the generator agent either: reaches the target, makes as many per-tile changes as would correspond to 100\% of the map, or reaches the limit on its number of steps (equal to the square of the number of tiles on the map). 

In this work, we focus on PCGRL's narrow representation, in which the agent chooses what to build on each tile in a (possibly random) sequence. This representation performs comparably and has fewer actions than others using additional navigation (turtle) or tile-coordinate (wide) actions.

As in the original PCGRL, the generator-agent observes a cropped, one-hot encoded view of the game board. Additional scalar inputs are concatenated with the agent's 2D observation (channel-wise), corresponding to the direction (-1, 0 or 1) of the target change along some metric (e.g., an increase in path length). The agent's reward is the amount by which the level has approached (or moved away from) target metrics since the previous step.

Let $\mathbf{s_t}$ be a vector representing the user-defined control metrics  (e.g. number of enemies, nearest enemy) at time $t$. At the beginning of the episode, a target/goal vector $g$ is assigned corresponding to desired metrics in the output level. This defines the generator's task, determining both its conditional observation and its reward. 

The conditional observation vector, $c=\sign(\mathbf{g} - \mathbf{s_t})$ represents the target directions for each metric. The loss of a given level with respect to the goal vector is given by $l_t = \abs(||\mathbf{g}-\mathbf{s_t}||_{L_1})$. Then the agent's reward at $t$ is $r_t = l_{t-1} - l_{t}$. The agent is rewarded for edits that close the gap between the level's metrics and the target, and punished for those that widen it.

To make training more efficient, control targets are sampled to maximize the agent's absolute learning progress \cite{portelas2020teacher}. If the generator has mastered some sub-space of targets perfectly, the control regime instead favors targets where the generator shows some long term change in performance (negative or positive). This allows the generator to focus on tasks that present the highest opportunity for improvement, as well as those that it may be forgetting.

While uniform sampling of targets is in theory enough incentive for an RL agent to learn perfect control (and it performs well in many of the minimal domains tested here), tasks in which large regions in the control space are impossible would cause a large compute overhead, with too much training time expended attempting to reach impossible targets. For example, generating a maze with the longest possible path but at the same time with an empty open area is impossible.

Generator agents are trained for 500 million frames using PPO \cite{schulman2017proximal} as implemented in stable-baselines \cite{stable-baselines}, and our content-generation tasks are implemented as gym environments \cite{gym}. Ranges for control targets were chosen based on estimated lower/upper bounds of these metrics on playable levels (e.g. we ask the generator to produce paths from anywhere between 0---no empty tiles on the board---and 112---the length of an optimal zig-zag path on a $14\times 14$ board).

\section{Results}

To visualize results, the generator is allowed to change the map up to 1,000 steps or until the usual termination conditions are met, on a random initial map for various control targets. To measure the agent's progress toward its goals and its diversity at each cell in the grid of control targets, the agent is similarly allowed to generate from a set of random initial maps. 

Diversity is computed as the mean per-tile hamming distance between the set of levels generated for a given target. The value is normalized using the maximum size of the map leading to values between 0 and 1, where 0 means that all the maps are the same, while 1 means that all the maps are 100\% different. Progress is defined as the relative percentage change from the initial state toward the control targets. For example, if the target path length is 20, the initial length is 10, and the agent manages to reach 15 by the end of the episode, then it has made 50\% ($\frac{15-10}{20 -10}$) progress toward its target. When visualizing results, we restrict mean progress to the interval $[0, 100]$, so that if the mean agent fails to move toward its target at all, or moves away from it, it is said to have made 0 progress.

\subsection{Original PCGRL environments}
\subsubsection{Binary}

\begin{figure}
    \centering
    \begin{subfigure}{0.5\textwidth}
    \centering
        \vspace*{1.5mm}
        \includegraphics[width=0.95\textwidth, trim=0 0 0 0, clip]{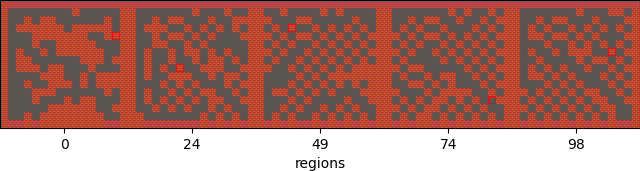}
        \par\medskip
        \includegraphics[width=0.98\textwidth, trim=10 40 0 0, clip]{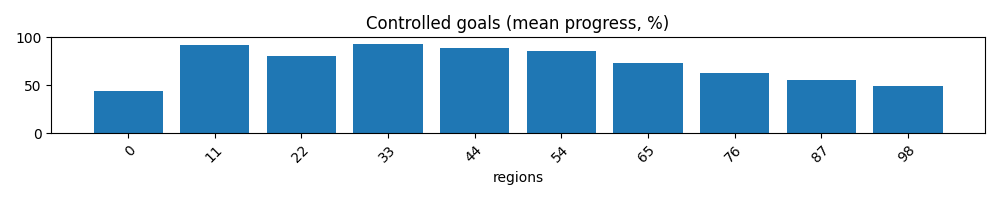}
        \includegraphics[width=0.98\textwidth, trim=5 0 0 0, clip]{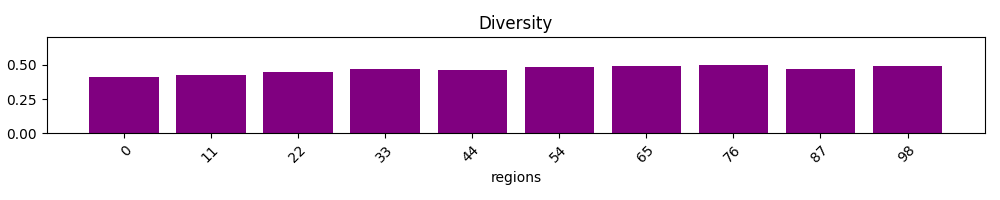}
    \end{subfigure}
    \caption{Generated examples from region controlled agent-generator. A border of checkerboard pattern is grown to efficiently increase the number of disconnected regions, with some open areas remaining at the center to contribute to path-length (a fixed goal, in this scenario).} 
    \label{fig:bin_regions}
\end{figure}

In the original PCGRL, generators are rewarded for creating levels with long paths and a single region. Here, we train a generator to control for one or both of these metrics. When we control for one metric, the other becomes a fixed goal (1 region or maximum path length) that is factored into the agent's reward function.

In the binary domain, generators learn precise control over the number of regions, while at the same time maximizing an (equally-weighted) reward for increasing path length (Figure~\ref{fig:bin_regions}). They balance a tightly packed checkerboard pattern with organic corridors of varying length and number. Maps become increasingly diverse as the number of regions increases and the path-length is forced to decrease, indicating that the generator has learned fewer maps with long paths than it has maps with medium or small paths overlaid with a checkerboard background. This is understandable, as there are more constraints on the form which a maximum-length path can take, as opposed to smaller paths which can for example be translated about the map to produce many distinct levels.

\begin{figure}
    \centering
    \begin{subfigure}{0.5\textwidth}
        \centering
        \vspace*{1.5mm}
        \includegraphics[width=0.95\textwidth, trim=0 0 0 0, clip]{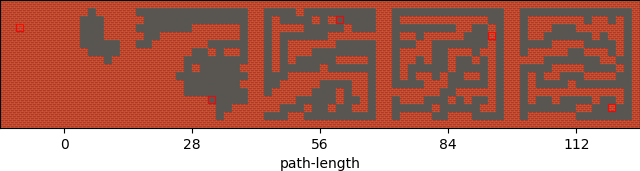}
        \par\medskip
        \includegraphics[width=0.98\textwidth, trim=10 45 0 0, clip]{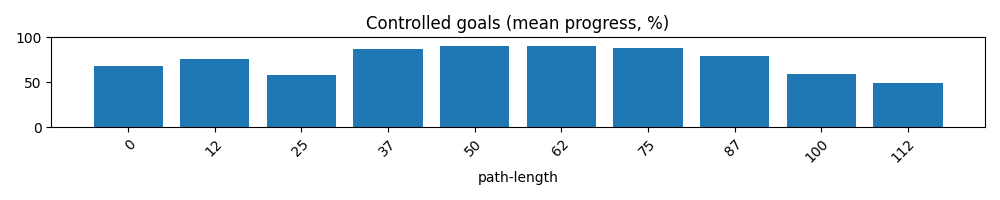}
        \includegraphics[width=0.98\textwidth, trim=5 0 0 0, clip]{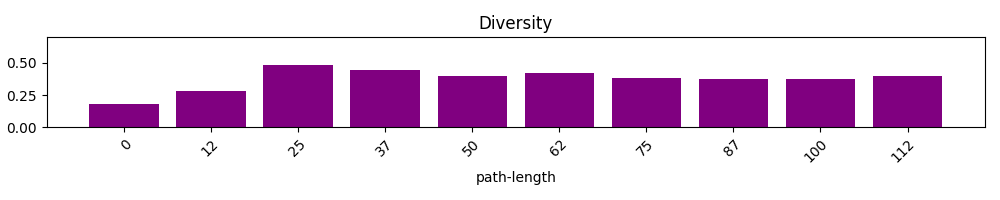}
    \end{subfigure}
    \caption{From left to right: the generator produces paths of increasing length. Its control weakens somewhat toward the longest path-lengths, which tend to be learned much later in training. There is surprising relative diversity among longer paths, compared to shorter ones which are less constrained and more numerous in theory.} 
    \label{fig:bin_path}
\end{figure}

Conversely, generators can control for specific path length while maintaining a single connected region, producing shorter paths in various forms and locations on the map, and a variety of organic labyrinths to achieve longer paths (Figure~\ref{fig:bin_path}). Again, diversity is highest among short path-length levels. During level-generation, path length tends to grow or shrink gradually, allowing the generator to sense whether it has reached or exceeded its target when its conditional input changes to 0 or changes sign, respectively.

\begin{figure}
    \begin{subfigure}{0.5\textwidth}
        \vspace*{1.5mm}
        \includegraphics[width=0.95\textwidth, trim=0 0 -30 0, clip]{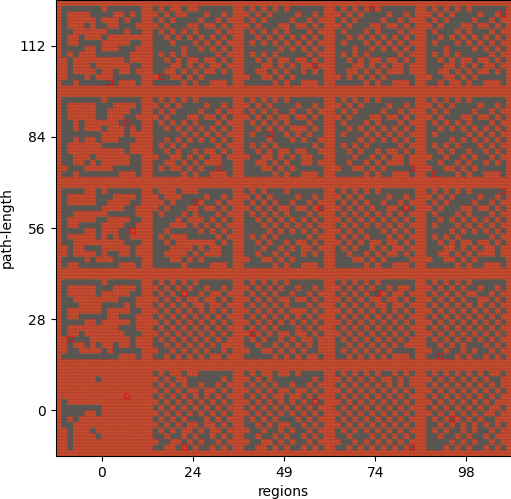}
    \end{subfigure}
        \begin{subfigure}{0.5\textwidth}
        \includegraphics[width=0.49\textwidth, trim=0 0 0 0, clip]{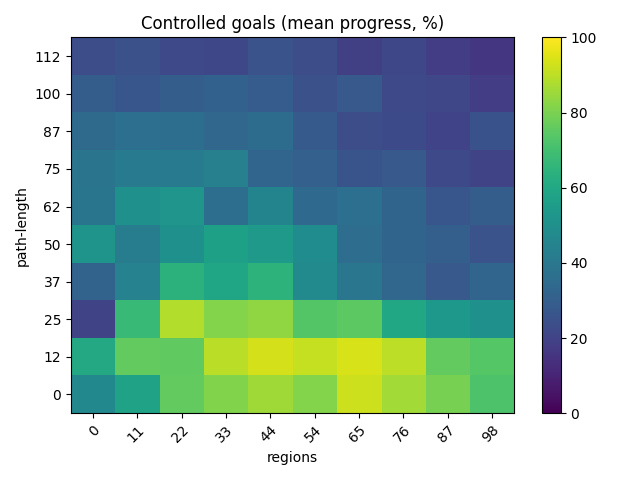}
        \includegraphics[width=0.49\textwidth, trim=0 0 0 0, clip]{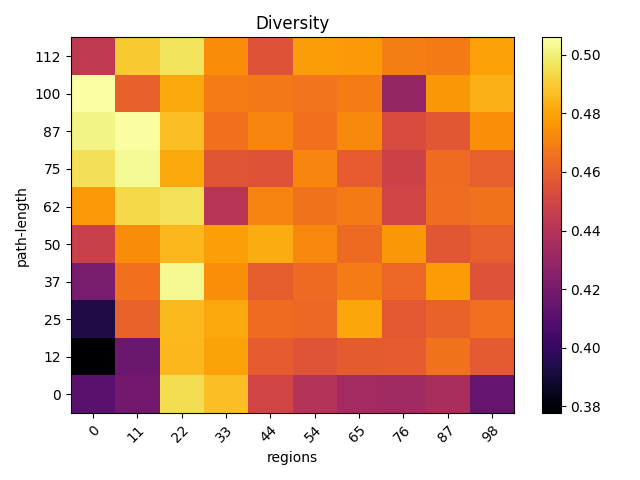}
    \end{subfigure}
        \caption{Generated levels produced by an agent-generator using varying path length and number of regions. It has precise control over the number of regions, but struggles to generate long paths. While it is practically impossible to reach many high path-length and high-regions targets, the agent is also less apt in generating long paths with few regions, likely because paths require more complex behavior to construct than atomic, disjoint regions.} 
    \label{fig:bin_regions_path}
\end{figure}

A generator trained to control both regions and path lengths learns to blend these checker-boarding and labyrinth-growing strategies (Figure~\ref{fig:bin_regions_path}).
It has the most success producing levels with low path-length and varying regions, and generates the most diverse levels when attempting to produce few regions and high path-length.

The generator learns regions more quickly than path length, presumably because the policy it learns to control regions is very simple (adding or removing ``checkerboard'' tiles depending on the target at a given time-step). Still, progress toward longer path lengths is apparent, particularly when the target number of regions is minimal (given that increasing regions and path length are conflicting goals). It is also possible that the weights assigned to these respective control metrics are such that it is often optimal for the generator, in terms of its reward, to sacrifice path-length to allow for more regions.

\subsubsection{Zelda}

In Zelda, agents are usually rewarded for placing 1 key, door, and player, between 2 and 5 enemies, and for creating a maximum path length (from the player to the key to the door) and a minimum distance between the player and the nearest enemy of at least 5. To train a controllable generator, we take nearest-enemy and path-length as our control metrics (as they could be taken as proxies for level difficulty), and treat as fixed any targets for which we are not controlling in a given experiment.

\begin{figure}
    \centering
    \begin{subfigure}{0.5\textwidth}
        \centering
        \vspace*{1.5mm}
        \includegraphics[width=0.95\textwidth, trim=0 0 0 0, clip]{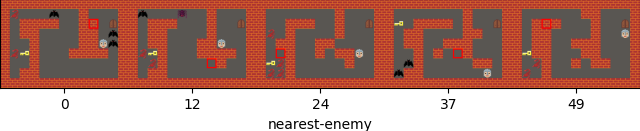}
        \par\medskip
        \includegraphics[width=0.98\textwidth, trim=10 40 0 0, clip]{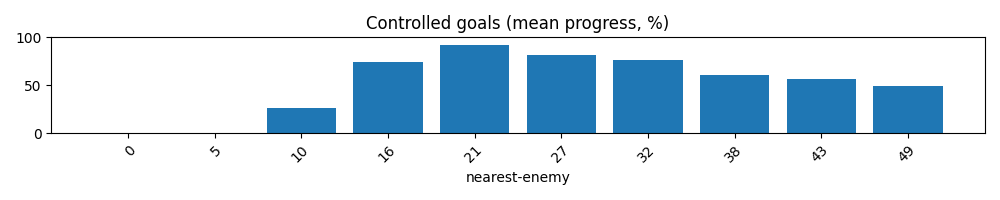}
        \includegraphics[width=0.98\textwidth, trim=5 0 0 0, clip]{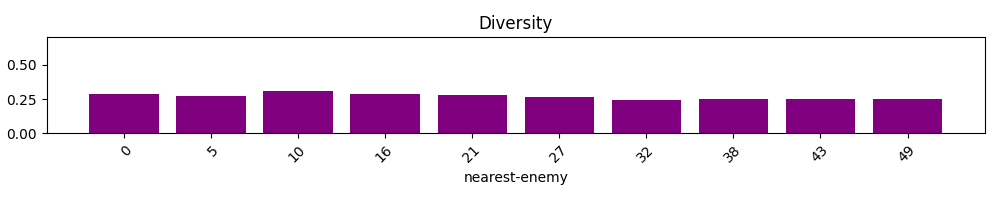}
    \end{subfigure}
    \caption{From left to right: the generator moves the nearest enemy further away from the player. It seeks to maximize the path length of the shortest possible solution in all cases. To this end, the generator relies on a repeated corridor structure, so diversity is greater when the nearest enemy is close, and other enemies may be placed anywhere further along the corridor.} 
    \label{fig:zelda_nearest}
\end{figure}

When controlling for nearest-enemy alone, the generator is able to maintain high path lengths while placing enemies at various minimum distances from the player (Figure \ref{fig:zelda_nearest}). These levels tend to share a common structure resulting in high path-length---namely a corridor through which Link must travel back and forth to retrieve the key and access the door---with enemies sliding up and down the corridor between levels to end up closer or further from the player. These levels are consequently more diverse when the closest enemy is nearer the player, since there is less constraint on the positioning of any other enemies in the level.

\begin{figure}
    \centering
    \begin{subfigure}{0.5\textwidth}
        \centering
        \vspace*{1.5mm}
        \includegraphics[width=0.95\textwidth, trim=0 0 0 0, clip]{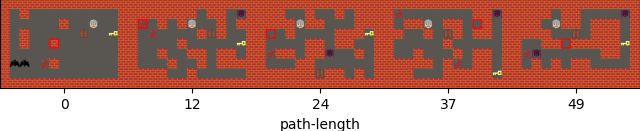}
        \par\medskip
        \includegraphics[width=0.98\textwidth, trim=10 40 0 0, clip]{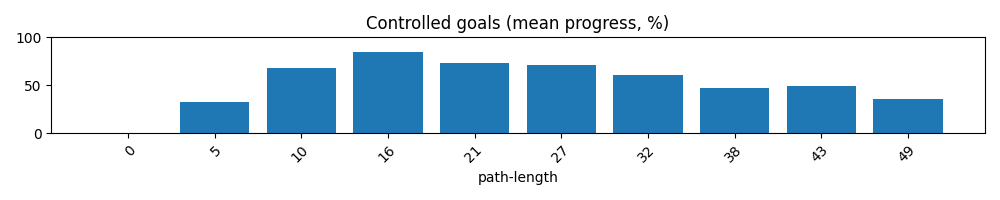}
        \includegraphics[width=0.98\textwidth, trim=5 0 0 0, clip]{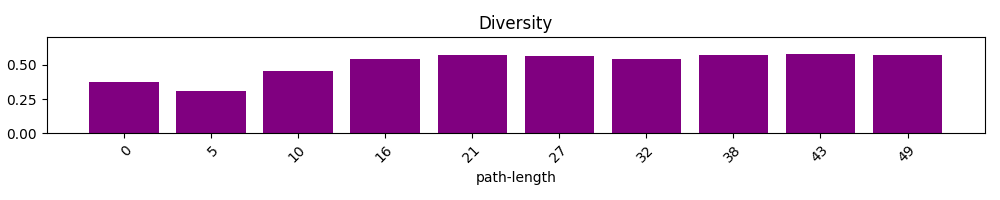}
    \end{subfigure}
    \caption{From left to right: the generator increases the path length of the shortest possible solution, keeping enemies at least 5 tiles from the player.} 
    \label{fig:zelda_path}
\end{figure}

Controllable path-lengths are produced by levels ranging from open spaces to more lengthy corridors, with the key and door placed at various distances from the player and each other throughout (Figure \ref{fig:zelda_path}). Surprisingly, there is greater diversity among higher path-length levels. 

\begin{figure}
\begin{subfigure}{0.5\textwidth}
    \vspace*{1.5mm}
    \includegraphics[width=0.95\textwidth, trim=0 0 -30 0, clip]{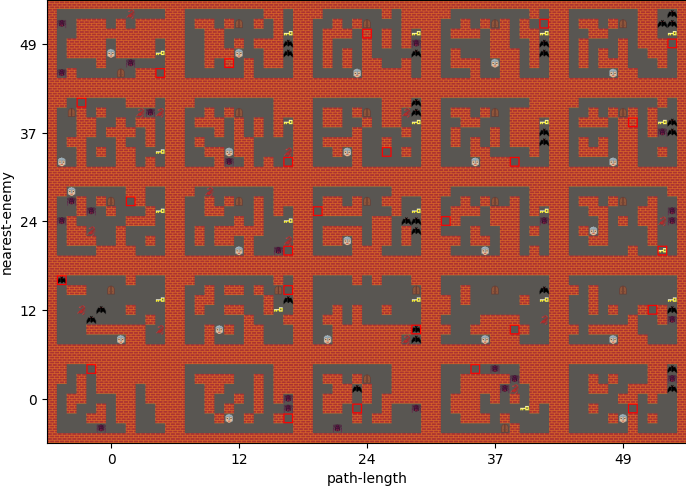}
    \par\medskip
    \includegraphics[width=0.49\textwidth]{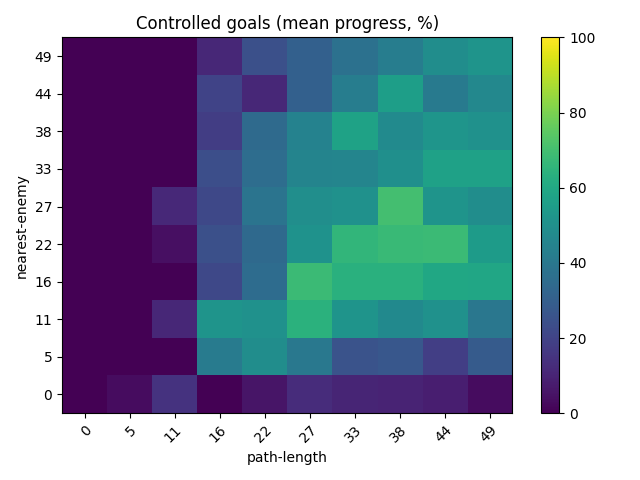}
    \includegraphics[width=0.49\textwidth]{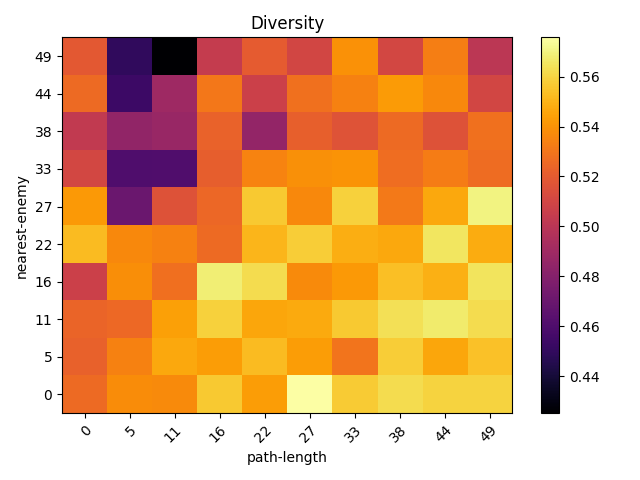}
\end{subfigure}
\caption{Left to right: generator increases path length. Bottom to top: generator increases distance to nearest enemy. The generator has the best control over a set of levels in which path-length is around twice nearest-enemy: enemies are placed by the key, with the door close to the player. Other targets are generally met reliably across the control-space.}
\label{fig:zelda_nearest_path}
\end{figure}

The generator can also be trained to control both metrics simultaneously (Figure \ref{fig:zelda_nearest_path}). In this case, it tends to favor levels in which the enemy is very close to either the player or the key/door.

\subsubsection{Sokoban}

In sokoban, the default heuristics define a good level as having 1 player, at least 2 crates, a matching number of targets, and a maximal solution length (trajectory of the player pushing crates onto targets). Here, the generator attempts to control the number of crates and solution-length.

The sokoban level-generator exhibits strong control over a range of solution lengths, corresponding to modest complexity on a very small map (Figure \ref{fig:sok_sol}). The generator reliably produces playable levels, and deviates from its control targets only when these contradict fixed playability targets. It produces a variety of levels, from simple open rooms, to simple obstacles, and corridors that require ``backtracking'' before moving the crate toward its target. However, most of its solutions use 1 crate in concert with player backtracking through narrow corridors.

\begin{figure}
    \centering
\begin{subfigure}{0.5\textwidth}
    \vspace*{1.5mm}
    \centering
    \includegraphics[width=0.95\textwidth, trim=0 0 0 0, clip]{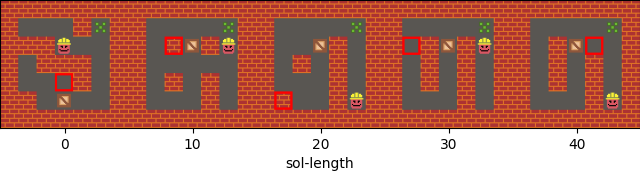}
    \par\medskip
    \includegraphics[width=0.98\textwidth, trim=10 40 0 0, clip]{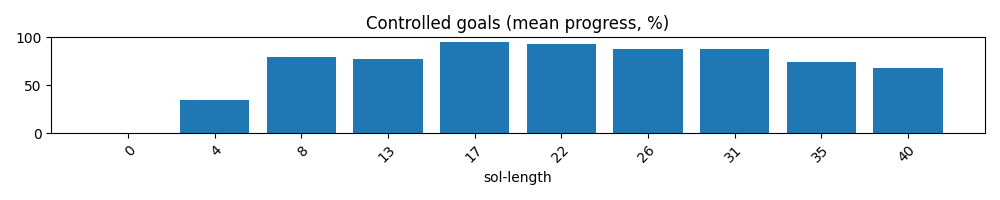}
    \includegraphics[width=0.98\textwidth, trim=4 0 0 0, clip]{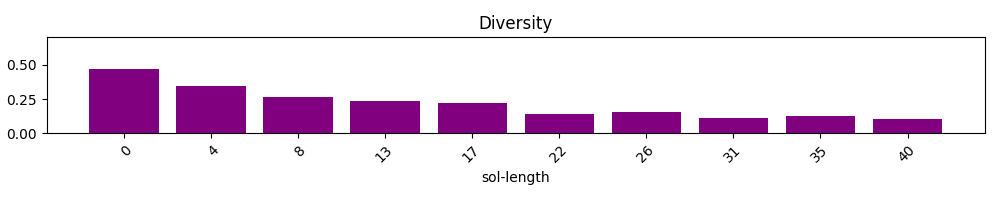}
\end{subfigure}
\caption{From left to right: the generator produces small sokoban levels with increasingly lengthy solutions. Longer solution lengths require the player to navigate obstacles and backtrack in order to push the crate to the target. Diversity decreases with level complexity.}
\label{fig:sok_sol}
\end{figure}

When asked to control the number of crates in addition to the solution length (Figure~\ref{fig:sok_crate_sol}), the generator learns to use multiple crates to increase level complexity instead of creating the kind of narrow corridors seen in Figure~\ref{fig:sok_sol}.
Low-crate levels are also more diverse compared to higher crate levels. This may be because far fewer solvable high-crate configurations can exist on a very small map, or because the generator has not discovered and honed in on optimal strategies, and is simply acting more randomly in this area of control space.

\subsection{micro-RCT (Roller Coaster Tycoon)}

Generators learn to build parks with specified levels of guest (un)happiness (Figure~\ref{fig:rct_happiness}). To maximize happiness, a cluster of concession stands is placed by the entrance, and guests immediately buy food/drinks whenever their hunger/thirst falls below a certain threshold. They then receive a happiness boost from the perceived value of the purchased item. To make guests unhappy, the generator places a few small thrill rides by the entrance. The rides are popular among guests, but have a high enough nausea score to induce some vomiting, which causes guests to become disgusted by their cramped surroundings.

\begin{figure}
\begin{subfigure}{0.5\textwidth}
    \vspace*{1.5mm}
    \includegraphics[width=0.95\textwidth,trim=0 0 -30 0, clip]{imgs/sokoban/narrow/alp_gmm/__sol-length_,__crate____3,_-bins_levels.png}
    \par\medskip
    \includegraphics[width=.49\textwidth]{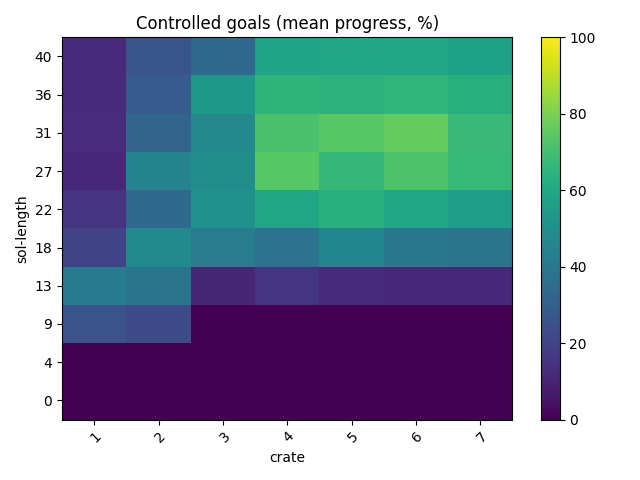}
    \includegraphics[width=.49\textwidth]{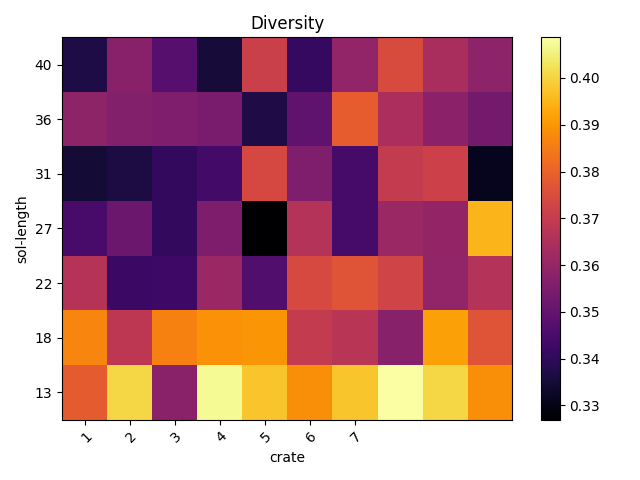}
\end{subfigure}
\caption{The generator produces small sokoban levels with various solution lengths and numbers of crates. It has the best control over levels with many crates and long solutions. Levels with few crates are the most diverse, as on such a small map, more crates drastically reduces the number of solvable configurations.}
\label{fig:sok_crate_sol}
\end{figure}

In micro-rct, guest happiness is sensitive, and small actions can tip the entire park's happiness over the edge (e.g. a thrill ride that causes a cycle of vomit). The generator mitigates this by acting dynamically (e.g. adding first aid stalls and deleting rides intermittently). The trained model struggles how to maintain a neutral level of happiness (which would be brought about by an empty park), likely because of this dynamic interaction.

\begin{figure}
    \centering
    \begin{subfigure}{0.5\textwidth}
        \vspace*{1.5mm}
        \centering
        \includegraphics[width=0.9\textwidth, trim=0 0 0 0, clip]{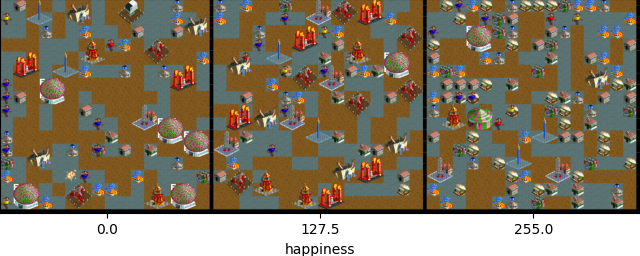}
        \par\medskip
        \includegraphics[width=0.94\textwidth, trim=10 45 0 0, clip]{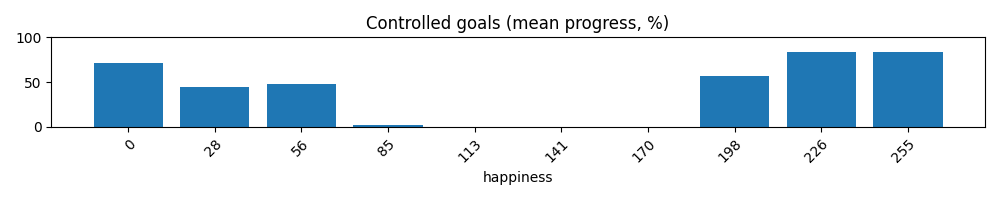}
        \includegraphics[width=0.94\textwidth, trim=5 0 0 0, clip]{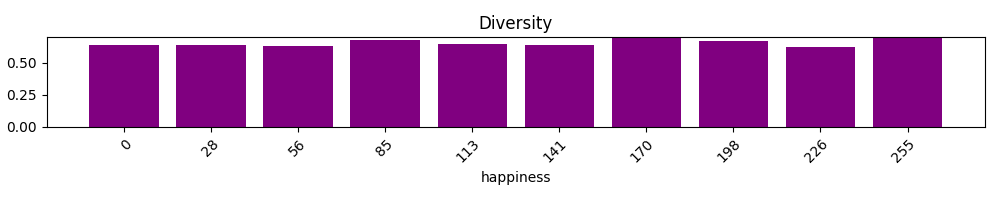}
    \end{subfigure}
\caption{From left to right: the generator produces minimal roller coaster theme parks of increasing guest happiness. Popular but nauseating thrill rides cause some vomiting and decrease happiness, while dense placement of burger stalls increases it. Extreme states of (un)-happiness are strong attractors. 
In medium happiness parks, the generator attempt to mitigate nausea-induced unhappiness by placing first-aid stalls throughout the park.
}
\label{fig:rct_happiness}
\end{figure}

\subsection{SimCity}

\begin{figure}
\centering
\begin{subfigure}{0.5\textwidth}
    \vspace*{1.5mm}
    \centering
    \includegraphics[width=0.3\textwidth,trim=330 0 630 0,clip]{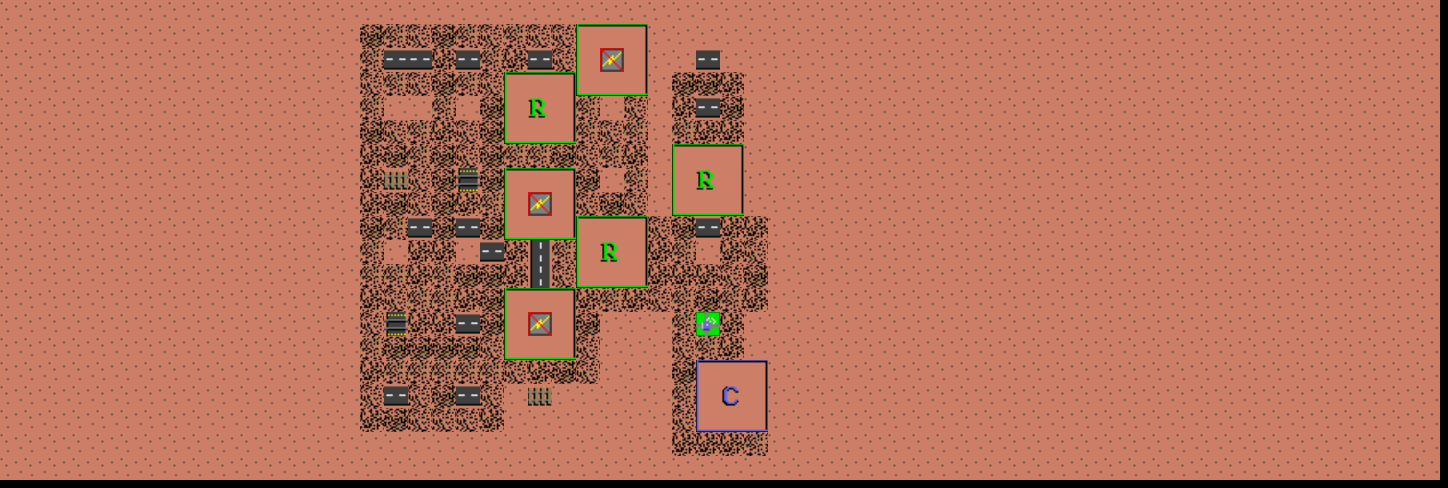}
    \includegraphics[width=0.3\textwidth,trim=330 0 630 0,clip]{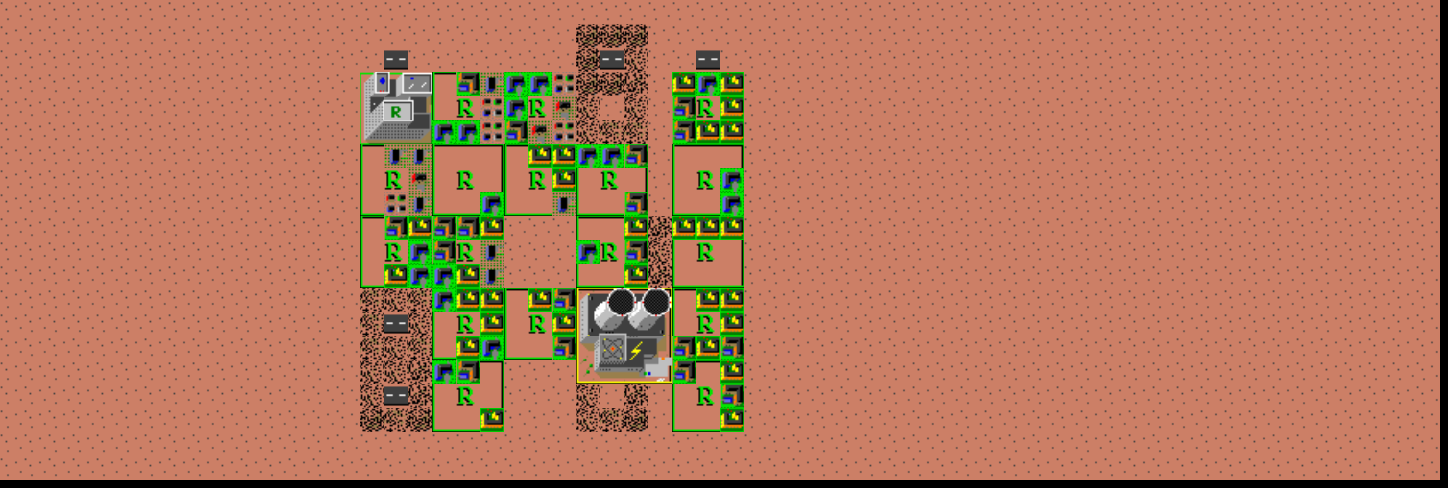}
    \includegraphics[width=0.3\textwidth,trim=330 0 630 0,clip]{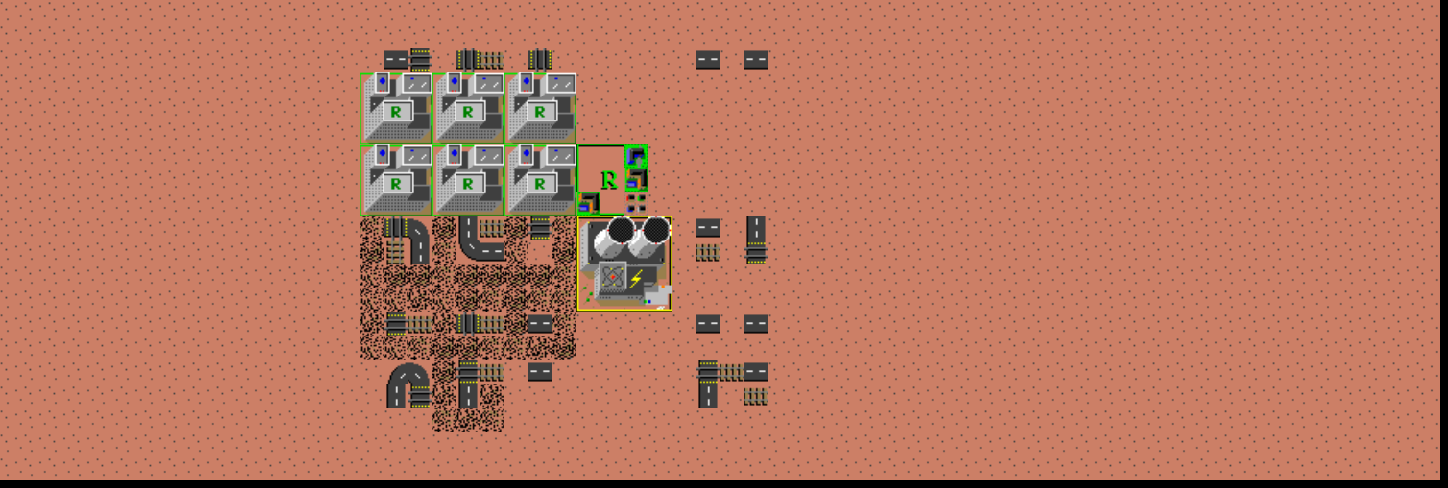}
    \par\medskip
    \includegraphics[width=0.94\textwidth, trim=10 45 0 0, clip]{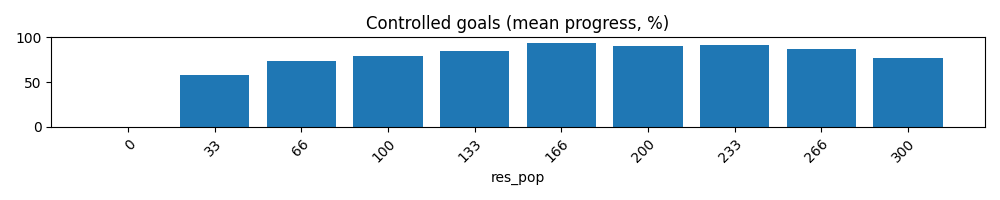}
    \includegraphics[width=0.94\textwidth, trim=6 0 0 0, clip]{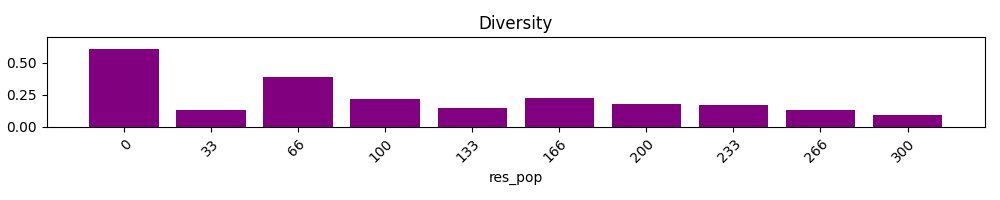}
\end{subfigure}
\caption{From left to right: the generator produces small SimCity layouts to control residential population. It connects zones to power by adjacency, and uses the presence of adjacent roads to toggle between low and high density to meet its target precisely. More populous city configurations tend to be less diverse, having been learned later in training and requiring more densely packed zones and roads.}
\label{fig:simcity_pop}
\end{figure}

In SimCity, the generator can learn to control the size of residential population in the city (Figure~\ref{fig:simcity_pop}). A maximal residential population can be achieved without any commercial or industrial zones via a strategic, dense placement of residential zones around a single power plant, with disjoint road tiles placed adjacent to zones to increase population density. The generator cycles through a wide diversity of unpopulated cities and settles on a handful of optimal configurations as target population increases.

That diversity is severely limited when the generator we evaluated attempts to produce populous cities likely reflects the fact that initial maps tend to have 0 population, and the agent learns higher-population layouts later in training. At the same time, the dense configurations leading to high population are necessarily less diverse than those allowing for unused space. This could be solved in future work by sampling different starting states involving partially-built cities, instead of starting from a completely empty city with zero population.

\section{Discussion}

Results on (the binary, zelda, and sokoban) level generation tasks show that with conditional (target) input and shaped reward, reinforcement learning can be used to train generators that encode a set of levels which are largely playable and controllably diverse along measures of interest, such as deterministic proxies for player difficulty. Experiments in management sim game-playing show that this learned control can handle the stochasticity and temporal dynamics resulting from agent-based simulation. These controls could serve as convenient and useful interfaces for game designers for producing content with given characteristics. They could also act as easily searchable or traversable spaces for the sake of curriculum generation, producing levels adapted to a particular player~\cite{yannakakis2011experience}, or for the purpose of training a player agent.

Notably, generators learn to adapt to changing, user-defined targets over the course of long inference episodes, despite having only been trained to approach one set of targets per episode. Novel states can emerge from this traversal of conditional space, and these dynamics set controllable content generators apart from methods that might search for a diverse set of levels directly, arguably giving them a greater degree of expressivity. On the other hand, the computation required to learn a generator is likely to be greater, since this diverse space of levels must be represented in a compressed form in the agent's weights, rather than stored in an archive.

Random initial states during level generation, combined with a limit on the number of changes that may be made to the map, ensure that a controllable agent learns diverse levels for targets where possible.
Often, the generator learns to traverse the space of levels in small steps, moving from initial states, incrementally along dimensions of interest until the target is met, thanks to its limited, binary conditional observation and its incentives for efficiency.
To ensure paths through this space are explored more thoroughly, and free them of the constraint of the initial random maps, an archive could be kept as in \cite{mordvintsev2020growing}.

In cases where there is some trade-off between control targets, the generator is tasked with finding optimal solutions according to some weighting of these controls' effect on reward, which may be difficult to tune. For example, in the binary problem, it is impossible to produce a maximal number of both paths and regions (since maximum path length implies 1 region, and maximal regions implies 1 path-length). It may be desirable instead to search for levels along the pareto front of these multiple objectives. This could be approximated by making additional controls of these per-control weights themselves, supplying them directly as conditional input, prompting the agent to learn multiple objectives.

\section{Conclusion}

Formulating content generation as a reinforcement learning problem allows us to learn generators that produce game levels of high quality along user-specified dimensions (e.g. path length). 
But it may also be desirable to have a generator that encodes the space these dimensions produce (e.g. levels of variable path length) rather than one point within it.

We show that this can be achieved by sampling from target points within this space over the course of training.
The generator then observes its target and is rewarded for approaching it.
The resulting learned generator is controllable, and the user is able to dynamically prompt it to produce levels of a certain type during inference.

To ensure that the reinforcement learning generator trains efficiently, for example by spending less time on large areas of level-space that are impossible to learn (e.g. high regions and path-length), we sample these targets to maximize the agent's absolute learning progress. The generator focuses on those areas where it is excelling or regressing the most, prompting it to explore new and disparate regions of level space while simultaneously recalling what it has already learned.  

When procedural content generation is used to learn better game AI, it may be desirable for the level generator to have an interpretable or easily searchable behavior space, bolstering our ability to manually or automatically create curriculums of levels for game-playing AI. 
The approach to learning controllable generators presented here is one such candidate, and our experiment suggest that it can learn to generate a diverse set of playable and complex levels.

The varying aptitude with which these controllable generators are able to explore the space of levels along user-specified dimensions can also help game designers to explore the constraints of level design in their game.
And the generator's ability to adapt to changing goals during inference, as well as the relative diversity of levels it can produce within a given set of target features, make it a potentially interesting co-creative tool.

\bibliographystyle{unsrt}
\bibliography{references}

\end{document}